\documentclass[letterpaper, 10 pt, journal, twoside]{IEEEtran}
\usepackage{amsmath,amsfonts}
\usepackage{algorithmic}
\usepackage{algorithm}
\usepackage{array}
\usepackage[caption=false,font=normalsize,labelfont=sf,textfont=sf]{subfig}
\usepackage{textcomp}
\usepackage{url}
\usepackage{verbatim}
\usepackage{graphicx}
\usepackage{cite}
\usepackage{enumerate}
\usepackage{float}
\usepackage{acro}
\usepackage{capt-of}
\usepackage{cuted}
\usepackage{stfloats}
\usepackage{soul}

\usepackage[font=small,labelfont=bf]{caption}

\usepackage{color}

\newcommand{\red}[1]{\textcolor{black}{#1}}

\graphicspath{{../}}

\DeclareAcronym{ADAS}{
    short = ADAS,
    long = Advanced Driver Assistance Systems
}
\DeclareAcronym{MOT}{
    short = MOT,
    long = Multi-Object Tracking
}
\DeclareAcronym{KF}{
    short = KF,
    long = Kalman Filter
}
\DeclareAcronym{SO}{
    short = SO,
    long = Sensor Object
}
\DeclareAcronym{SPENT}{
    short = SPENT,
    long = Single-Prediction Network
}
\DeclareAcronym{SANT}{
    short = SANT,
    long = Single-Association Network
}
\DeclareAcronym{MANTa}{
    short = MANTa,
    long = Multi-Association Network
}
\DeclareAcronym{RNN}{
    short = RNN,
    long = Recurrent Neural Network
}
\DeclareAcronym{LSTM}{
    short = LSTM,
    long = Long Short-Term Memory
}
\DeclareAcronym{BILSTM}{
    short = BILSTM,
    long = Bidirectional Long Short-Term Memory
}
\DeclareAcronym{GRU}{
    short = GRU,
    long = Gated Recurrent Unit
}
\DeclareAcronym{GT}{
    short = GT,
    long = Ground Truth
}
\DeclareAcronym{GNN}{
    short = GNN,
    long = Global Nearest Neighbor
}
\DeclareAcronym{JPDA}{
    short = JPDA,
    long = Joint Probabilistic Data Association
}
\DeclareAcronym{HA}{
    short = HA,
    long = Hungarian Algorithm
}
\DeclareAcronym{SoDA}{
    short = SoDA,
    long = Soft Data Association
}
\DeclareAcronym{FC}{
    short = FC,
    long = Fully Connected
}
\DeclareAcronym{MSE}{
    short = MSE,
    long = Mean Squared Error
}
\DeclareAcronym{ML}{
    short = ML,
    long = Machine Learning
}
\DeclareAcronym{NN}{
    short = NN,
    long  = Neural Network
}
\DeclareAcronym{TbD}{
    short = TbD,
    long  = Tracking-by-Detection
}
\DeclareAcronym{RMSE}{
    short = RMSE,
    long  = Root Mean Square Error
}

\begin{document}
\title{Tiny Neural Networks for Multi-Object Tracking in a Modular Kalman Framework}

\author{Christian Holz\textsuperscript{1}, Christian Bader\textsuperscript{1}, Markus Enzweiler\textsuperscript{2}, Matthias Drüppel\textsuperscript{3*}

\begin{small}
\textsuperscript{1}Daimler Truck AG, Research and Advanced Development, Stuttgart, Germany \\
\textsuperscript{2}Institute for Intelligent Systems, University of Applied Sciences, Esslingen, Germany \\
\textsuperscript{3}Center for Artificial Intelligence, Baden-Württemberg Cooperative State University (DHBW), Stuttgart, Germany \\
\textsuperscript{*}Corresponding author: matthias.drueppel@dhbw-stuttgart.de
\end{small}

}

\maketitle

\begin{abstract}

We present a modular, production-ready approach that integrates compact \ac{NN} into a Kalman-filter-based \red{Multi-Object Tracking} (MOT) pipeline. We design three tiny task-specific networks to retain modularity, interpretability and real-time suitability for embedded Automotive Driver Assistance Systems:
\begin{enumerate}[(i)]
\item SPENT (Single-Prediction Network) — predicts per-track states and replaces heuristic motion models used by the \ac{KF}.
\item  SANT (Single-Association Network) — assigns a single incoming sensor object to existing tracks, without relying on heuristic distance and association metrics.
\item MANTa (Multi-Association Network) — jointly associates multiple sensor objects to multiple tracks in a single step.
\end{enumerate}
Each module has less than 50k trainable parameters. Furthermore, all three can be operated in real-time, are trained from tracking data, and expose modular interfaces so they can be integrated with standard Kalman-filter state updates and track management. This makes them drop-in compatible with many existing trackers. \red{Modularity is ensured, as each network can be trained and evaluated independently of the others.} Our evaluation on the KITTI tracking benchmark shows that SPENT reduces prediction RMSE by more than 50\% compared to a standard Kalman filter, while SANT and MANTa achieve up to 95\% assignment accuracy. These results demonstrate that small, task-specific neural modules can substantially improve tracking accuracy and robustness without sacrificing modularity, interpretability, or the real-time constraints required for automotive deployment.
\end{abstract}

\renewcommand{\IEEEkeywordsname}{Keywords}
\begin{IEEEkeywords}
\normalfont Multi-Object Tracking, Recurrent Neural Networks, Kalman Filter, Real-Time Embedded Systems, Tiny Neural Networks, Data-Driven Methods
\end{IEEEkeywords}

\section{Introduction}
\IEEEPARstart{T}{he}
ongoing evolution of \ac{ADAS} has brought the need for precise and reliable \ac{MOT} into the spotlight {\cite{KF_Bewley.2016,KF_simple_cues.2022,KF_BoundingBD.2023, DL_RNN_mot.2016,DL_RNN_data_association.2019,DL_CNN_ATT_mot.2017,DL_ATT_Trackformer.2022, DL_ATT_CNN_soda.2020, Zhang2021ByteTrackMT,Zhang2020FairMOTOT, liu2020deepmtt}}.
In complex and dynamic environments, as encountered in urban traffic, it is crucial to accurately capture and predict the positions of multiple objects already in the \red{early timestamps after detection} -- a key challenge in assisted and automated driving.
In the commonly used \ac{TbD} paradigm, a tracker fuses detected \red{\acp{SO}} to create consistent object tracks over time.
A crucial step within this paradigm is the association of the incoming measured \ac{SO} with their corresponding existing object tracks to update their properties.
If no association can be made, new object tracks must be initialized {\cite{KF_Bewley.2016,KF_simple_cues.2022,KF_BoundingBD.2023,OD_kampker.2018,OD_wu.2022}}.
Tracking frameworks form the heart of ADAS that are used in millions of vehicles around the globe.
The vast majority of these frameworks rely on classical approaches such as the \ac{KF} or its variants {\cite{KF_Bewley.2016,KF_simple_cues.2022,KF_BoundingBD.2023}}.
These classical tracking theories have the great benefit of being modular and interpretable.
The task is split into clearly separated subtasks such as the prediction of currently tracked objects and the association with newly measured ones.
However, development for automated driving is highly complex~{\cite{AD_Overview.2017,AD_book.2024}}.

\begin{figure}[!bp]
	\centerline{\includegraphics[width=0.95\linewidth]{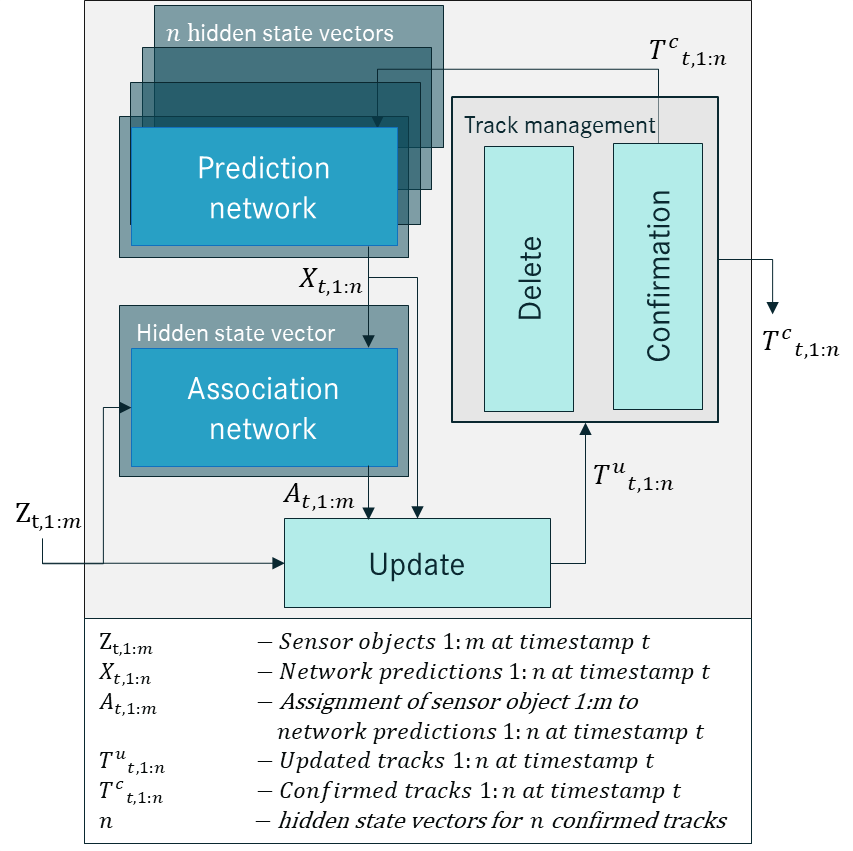}}
	\caption{
This schematic representation shows the integration of two \acp{NN} (highlighted in dark blue) within a TbD framework.
The "Association network" can be implemented using either \ac{SANT} or \ac{MANTa}. It works in tandem with the prediction network \ac{SPENT}, which takes tracked objects ${T^c}_{t,1:n}$ and predicts them to the next timestamp as $X_{t,1:n}$. The predicted objects are then associated with sensor observations $Z_{t,1:m}$ by \ac{SANT} or \ac{MANTa}.
}
	\label{pic::MotFramework}
\end{figure}
In the automotive industry, tracking systems are typically developed through a software platform that is designed to support a range of vehicle models, each of which will have varying sensor placements, system configurations, or even entirely different sensor suites. These systems must perform reliably across a wide spectrum of driving scenarios, which can introduce performance challenges in specific situations.
Traditional solutions often rely on heuristics and manual parameter tuning, making the software cumbersome to maintain and difficult to extend.
Moreover, these hand-engineered methods lack automated optimization, resulting in suboptimal performance in complex driving conditions. To address these limitations, we propose a data-driven tracking framework that allows for fine-tuning for specific configurations, thereby improving both maintainability and adaptability.

\section{Contributions and overview}
Our primary contribution is the development, \red{individual evaluation} and joint integration of three novel \ac{NN} that we label:
\begin{enumerate}[(i)]
    \item SPENT (Single-Prediction Network) which predicts the states of tracked objects.
    \item SANT (Single-Association-Network) which associates one incoming sensor objects to all currently tracked objects.
    \item MANTa (Multi-Association Network) which associates multiple incoming sensor objects to all currently tracked objects.
\end{enumerate}

These networks were specifically designed for real-time, embedded inference \red{{\cite{DL_TinyML.2019}}}. Each of them has fewer than 50k trainable parameters. Fig. {\ref{pic::MotFramework}} provides an overview of our approach in which the association network can be implemented using either \ac{SANT} or \ac{MANTa}.
The input for the proposed prediction network (i) \red{is up to $m$ \acp{SO}} $Z_{t,1:m}$ at timestamp $t$.
If new objects are detected by the sensors, these are stored in a $k$-dimensional state vector containing information such as object position $(x,y)$, yaw angle and object dimensions
(length and width) (for $k=5$).
\ac{SPENT} predicts all currently tracked objects $X_{t,1:n}$ (up to $n$) to the next timestamp, \red{where} they are used as input to either the SANT or MANTa association networks.
These provide the association matrix $A_{t,1:m}$, that is used to update the tracks ${T^u}_{t,1:n}$ using the corresponding sensor objects \red{or to create new tracked objects.}
\red{The Track Management can then decide to delete tracks that were not updated for a specific amount of time, and send out tracks ${T^c}_{t,1:n}$ to the next higher software component when they have been confirmed by sensor objects.}
(i) In contrast to the \ac{KF} {\cite{KF_BoundingBD.2023}}, our proposed prediction network is capable of predicting the state of individual objects without the need for a predefined heuristic prediction model. 
The self-learning, data-driven approach enables adaptability to various scenarios and the ability to effectively handle non-linearities and \red{behaviors} of road users.
Many conventional tracking systems rely on static methods for data association.
Commonly used algorithms like the \ac{HA} {\cite{DA_hungarian.1955}} require heuristics and fixed thresholds.
(ii) Our proposed \ac{SANT} and (iii) \ac{MANTa} replace the calculation of a distance metric for the corresponding assignment by employing \ac{ML} in order to resolve situations unclear for traditional approaches.
\red{Both our} prediction network and our association networks can be developed and evaluated as stand-alone models.
For a \red{thorough} evaluation, we proceed by integrating them into an existing tracking system and demonstrate their performance through multiple tests and comparisons with established methods.
This work provides new insights and advancement in the development of \ac{ADAS} tracking systems by applying \ac{ML}\@.

\section{Related work}
\subsection{State prediction for tracked objects}
One fundamental problem in \ac{TbD} frameworks is the prediction of the states of the already tracked objects.
\red{In many approaches, Kalman filters} and their variants have proven to be effective for state prediction {\cite{KF_simple_cues.2022, KF_Bewley.2016, KF_framework.2013}}.
However, they reach their limits in more complex scenarios, particularly in the presence of non-linear motion patterns and interactions among multiple objects {\cite{KF_Ristic.2004,KF_Julier.2004,KF_Wan.2000}}.
Ristic et al. {\cite{KF_Ristic.2004}} highlight the limitations of Kalman Filters in handling nonlinear and non-Gaussian cases, introducing Particle Filters as a potential alternative.
Julier et al. {\cite{KF_Julier.2004}} extend the standard Kalman Filter with the Unscented Kalman Filter (UKF) to better address nonlinear motion models.
Wan et al. {\cite{KF_Wan.2000}} propose the use of Gaussian Mixture Models for tracking multiple objects in cluttered environments.
In this work, we introduce a novel \ac{MOT} approach that leverages \ac{ML} to overcome these challenges.
We specifically focus on the development and implementation of \acp{NN}, which can enable more precise and flexible data-driven object state predictions.

\subsection{Association of sensor objects to tracks}
Another key challenge for TbD trackers is data association.
The widely used \ac{GNN} algorithm, often implemented via the \ac{HA} {\cite{DA_hungarian.1955}}, assigns detections to tracks by minimizing a distance metric.
However, it only considers current observations, ignoring temporal continuity and motion patterns, which can lead to errors in complex scenarios such as crossing objects or noisy sensor data.
\red{Practical detection-centric association strategies (e.g., CenterTrack, TransTrack) pursue robust, detection-driven matching across frames and have shown strong empirical performance {\cite{Zhou2020CenterTrack,Sun2020TransTrack}}.}
While methods like the \ac{JPDA} {\cite{DA_JPDA.1993}} evaluate the likelihood of all possible assignments, they are computationally expensive.

\ac{ML}-based approaches have been proposed to address these limitations.
The temporal information in tracks can for example be leveraged through the attention mechanisms as used in {\cite{DL_ATT_CNN_soda.2020, DL_ATT_CNN_mot_sot_based.2017}} or \acp{RNN} as developed in {\cite{DL_RNN_mot.2016, DL_RNN_data_association.2019}}.
The latter is what we are also pursuing in this work.
Similar to the problem statement by Mertz et al. {\cite{DL_RNN_data_association.2019}}, the aim of our work is to develop a data-based approach that can learn to completely solve the combinatorial non deterministic polynomial time (NP) hard optimization problem of data association.
In the context of this work, we put forward the hypothesis that a \ac{GRU}-based association network can be designed and trained \red{without a predefined distance measure}, as discussed in the next chapter.

\subsection{Real time applications}
Tracking systems for ADAS run on embedded devices in real-time. They must \red{provide immediate} state prediction of objects directly after their first detection {\cite{KF_Bewley.2016,DL_Wojke.2017}}, making offline tracking methods like {\cite{offline_mot.2017}} unsuitable. Consequently, modern multi-object tracking approaches rely on online methods, which do not have access to future sensor data. These methods estimate object-track similarity based on predicted positions or object features like appearance.
\red{Recent works focusing on real-time design choices highlight trade-offs between accuracy and computational cost relevant to embedded applications {\cite{Wang2019RealtimeMOT,Zhou2020CenterTrack}}.}

\paragraph{Kalman Filter based tracker}
Our \ac{ML} models are integrated into a \ac{KF}-based tracking system, widely used for its robust performance and interpretability {\cite{KF_Wan.2000,KF_Ristic.2004,KF_Julier.2004,KF_framework.2013,KF_Bewley.2016,KF_simple_cues.2022,KF_BoundingBD.2023}}. Bewley et al. {\cite{KF_Bewley.2016}} introduced an efficient multi-object tracking method combining a KF with the Hungarian Algorithm. Seidenschwarz et al. {\cite{KF_simple_cues.2022}} proposed a simpler approach based on visual cues like color, shape, and motion for object tracking and frame-to-frame association, avoiding the complexity of many modern trackers.

\paragraph{Recurrent Neural Network based Tracker}
Similar to our methodologies, RNN-based approaches for online multi-target tracking have been introduced in~{\cite{DL_RNN_mot.2016,DL_RNN_data_association.2019,DL_RNN_data_association.2020,liu2020deepmtt}}.
Specifically, the work by Mertz et al. {\cite{DL_RNN_data_association.2019}} focuses on data association within a TbD framework.
Their proposed DeepDA model, a Long Short-Term Memory (LSTM)-based Deep Data Association (DeepDA) Network, is designed to learn and execute the task of associating objects across frames.
This model's ability to discern association patterns directly from data enables a robust and reliable tracking outcome, even in environments with significant disturbances.
Mertz et al. employ a distance matrix, derived from the Euclidean distance measure, as the input for the DeepDA network.
This innovative approach effectively supersedes traditional association algorithms, such as the Hungarian Algorithm.
It is inferred that the Euclidean distance measure served as a foundation not only for generating the ground truth (GT) training data (i.e., distance matrices) but also for the subsequent evaluation process.
However, this is not explicitly stated.
In our work, we want to enable the network to follow a completely data-driven association logic without forcing a concrete distance metric.

\paragraph{Attention Mechanism based Tracker}
In this paper, we analyze tracks using \ac{LSTM}-based models.
An alternative architecture would be the attention mechanism {\cite{ML_Attention.2017}}, utilized in various studies {\cite{DL_ATT_mot_sot_based.2017,DL_ATT_CNN_mot_sot_based.2017,DL_ATT_CNN_soda.2020,DL_CNN_ATT_mot.2017,DL_ATT_Trackformer.2022,Carion2020EndtoEndOD}}.
Hung et al. {\cite{DL_ATT_CNN_soda.2020}} focus on soft data association (SoDA), enabling probabilistic associations and accounting for uncertainties by aggregating information from all detections within a temporal window.
This approach allows the model to learn long-term, interactive relationships from large datasets without complex heuristics.
However, since tracking tasks involve real-time data processing with relatively short sequences, \acp{RNN} can efficiently handle this without the overhead of calculating attention weights for every input, making them more computationally efficient for short to medium-length sequences.

\section{Tracking with \ac{ML}-based Prediction and Association Networks}
We apply the \ac{TbD} paradigm, in which a tracker fuses object detections to generate object tracks that are consistent over time. In our study a \ac{KF} framework was implemented following the computational ideas of Vo et al. {\cite{KF_framework.2013}}.
To enable our models to incorporate temporal information, we use LSTM {\cite{DL_LSTM.1997}} and bidirectional Long Short-Term Memory (BILSTM) network layers {\cite{DL_BILSTM.1997}}, both for the prediction and the association of the sensor objects to the existing tracks.

\subsection{Single Prediction Network (SPENT)}
\label{ssec:spent}
Our approach uses the hidden states of the \ac{LSTM} layer as an information repository for each object. {\ac{SPENT}} operates in an open-loop manner, predicting future states based on past data.
This allows the network to predict the most likely state of an object for the next timestamp.

\paragraph{Data preprocessing}
In the development of our model, \ac{GT} data comprising vehicle tracks (cars and vans) from the KITTI dataset {\cite{Geiger2012CVPR}} was utilized.
We extracted 635 tracks, filtering out shorter tracks using a 3-frame threshold, resulting in 624 tracks.
This ensures the network receives tracks with sufficient timestamps, a common practice in tracking systems \cite{DL_DetTrack.2023}.
The track lengths \red{vary} from a minimum of 4 frames to a maximum of 643 frames (Sequence 20, ID 12).

\begin{figure}[b!]
	\centerline{\includegraphics[width=0.91\linewidth]{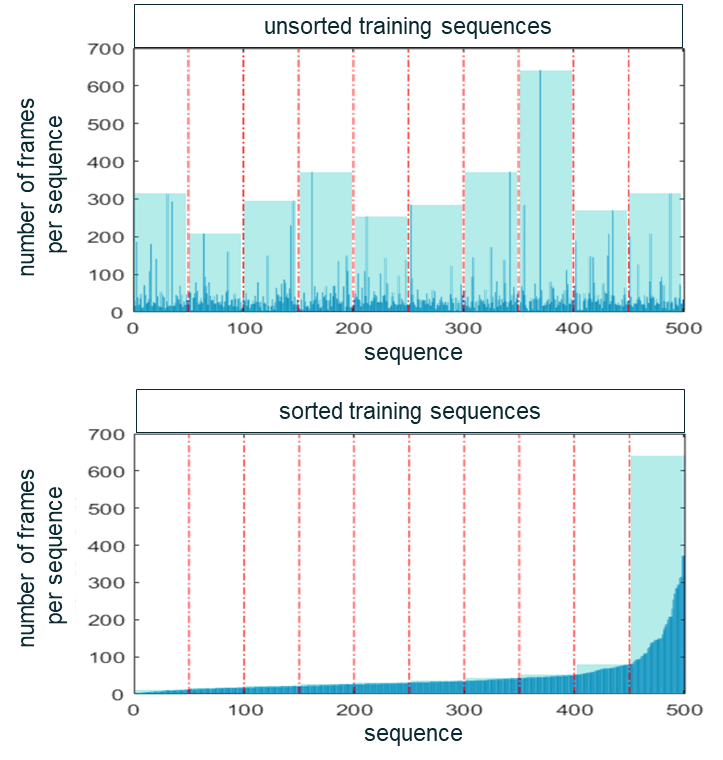}}
    \caption{Analysis of sequence padding: unsorted vs. sorted data. This figure illustrates the impact of sequence padding on LSTM training based on the sorting of input data. The upper panel shows that unsorted data requires extensive padding to equalize batch sequence lengths, increasing computational overhead. In contrast, the lower panel demonstrates that sorting data by length before batching significantly reduces the necessary padding.}
	\label{pic::Padding}
\end{figure}

To enhance model generalization and foster convergence during training, we normalized the state values of tracks at time $t$ (predictors) and at time $t+1$ (targets) in accordance with the methodology outlined in {\cite{DL_book.2019}}.
This normalization process standardizes the distribution of both predictors and targets to have a mean of zero and a unit variance.
The mean values $\vec{\mu}$ and standard deviations $\vec{\sigma}$ for each state in the state vectors $\vec{Z}_t$ were computed across all tracks. For this, all tracks were joined together, leading to one pseudo-track with a total number of timestamps $N$:

\begin{equation}
	\vec{\mu} = \frac{1}{N} \sum_{t = 1}^{N} \vec{Z}_t
\quad\text{and}\quad
	\vec{\sigma} = \sqrt{{\frac{1}{N - 1}} \sum_{t = 1}^{N} (\vec{Z}_t - \vec{\mu})^{2^*}},
\end{equation}
\noindent
\red{where the square $(...)^{2^*}$} and the square-root must be applied element wise.

\red{In our approach, we apply pre-padding (padding at the beginning of sequences) as described by Reddy et al. {\cite{DL_padding.2019}}, who examined the impact of padding strategies on sequence-based \acp{NN}.
They show that while both pre-padding and post-padding are possible, the choice affects how sequence context is preserved in \ac{LSTM}-based networks: pre-padding keeps recent timestamps aligned to the sequence end, which can be beneficial for certain recurrent architectures.
To manage varying sequence lengths, we pad shorter sequences with tokens placed at the start of the sequence (pre-padding), ensuring all sequences in a batch have uniform length.
We use zeros as padding tokens inserted at the beginning of sequences as needed; this enables efficient batch processing while keeping the most recent (non-padded) data aligned at the sequence end.
As noted by Reddy et al. {\cite{DL_padding.2019}}, padding introduces noise, but it is necessary for aligning sequences in mini-batches for \ac{LSTM} training. To reduce the amount of padding (and therefore the introduced noise), we sort the training dataset by sequence length before creating mini-batches and applying the pre-padding. This method, illustrated in Fig. {\ref{pic::Padding}}, significantly minimizes the padding (shown in turquoise) required for each mini-batch and is important for achieving convergence during training.}

\begin{figure}[!ht]
	\centerline{\includegraphics[width=0.95\linewidth]{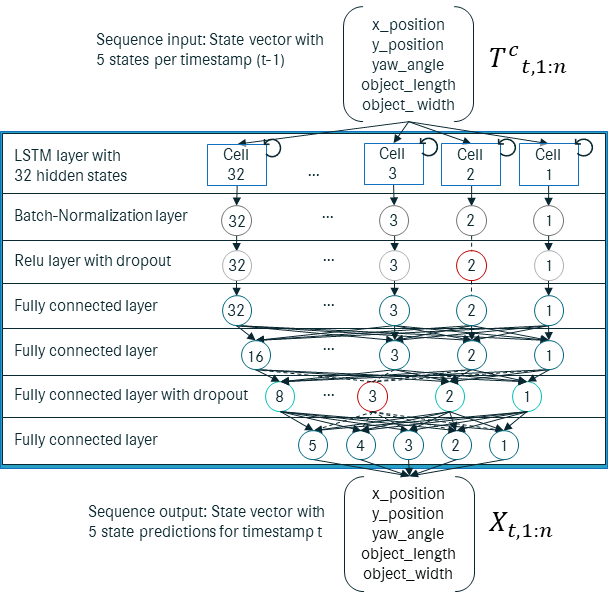}}
	\caption{Schematic representation of the generic structure of \ac{SPENT}. \red{SPENT replaces the classical motion model of a Kalman filter by directly predicting the next states estimate $X_{t}$ from the current confirmed track states $T^c_{t}$. SPENT outputs the predicted state mean only, while uncertainty propagation (state covariance) and measurement updates remain within the classical Kalman filter framework. This design ensures drop-in compatibility with standard Kalman-based Tracking-by-Detection pipelines.}}
	\label{pic::SpentArch}
\end{figure}

\paragraph{Network architecture}
As depicted in Fig. \ref{pic::SpentArch}, the schematic illustration of the generic structure of the prediction network illustrates how the architecture is adeptly designed to address the challenges of real-time state prediction.
This representation highlights the strategic deployment of the \ac{LSTM} layer for storing and processing object-specific information, facilitating accurate and timely predictions of object states.

The foundational layer of our \ac{SPENT} is an \ac{LSTM} layer, where hidden states are dynamically updated at each timestamp based on incoming measurement data.
This mechanism allows continuous correction throughout each track's sequence, enhancing predictive accuracy.
The number of hidden units correlates with the amount of information retained over time, as shown in Fig. \ref{pic::SpentArch}.
These hidden states encapsulate information from all preceding timestamps, ensuring comprehensive temporal understanding.
Following the \ac{LSTM} layer, we incorporate a Batch-Normalization layer which accelerates training and promotes convergence by mitigating internal covariate shift \cite{DL_batch_norm.2015}.
Next \red{is} a \red{ReLU} layer, which applies a non-linear threshold operation, setting values below zero to zero.
During training, a Dropout layer randomly nullifies input elements with a specified probability, regularizing the model and preventing overfitting \cite{DL_dropout.2014}.
The architecture concludes with a Fully Connected (FC) layer, which integrates insights from previous layers, with its dimensionality aligned to the number of required output variables {\cite{DL_FC.2010}}.
Our model’s loss function is based on the Mean Squared Error (MSE) metric, which is calculated for each state value prediction.
The MSE quantifies the average squared discrepancy between the predicted and actual target values.
We chose MSE because it provides a clear and direct measure of how closely our predictions align with the true states, making it an effective metric for optimizing our model's state predictions.

For one single prediction the \ac{MSE} is given by
\begin{equation}
	MSE = \frac{1}{k}\sum_{i=1}^{k}{({Z}_i-{{X}}_i)}^2,
\end{equation}
\noindent
\red{where $k$ is the} length of the predicted state vector (here $k=5$), $Z_i$ are the entries of the ground truth state vector (KITTI cars and vans tracks) and ${X}_i$ the respective entries of the predicted state vector from our network.
During training, the cost function is evaluated for one mini-batch with several sequences and a total number of $N$ timestamps.
It is calculated as half the mean-square-error of the predictions added up for each timestamp, normalized over all timestamps.
The factor of $\frac{1}{2}$ simplifies the gradient during backpropagation:
\begin{equation}
	\red{cost = \frac{1}{2}\frac{1}{N} \sum_{t=1}^{N}\frac{1}{k}\sum_{i=1}^{k}{({Z_{t,i}}-{{X}_{t,i}})}^2},
\end{equation}
\noindent
\red{where $Z_{t,i}$ refers} to the $i$-th entry in the state vector at timestamp $t$.

\paragraph{\red{Overall training setup}}
\red{Our \ac{LSTM} layers use tanh activation for cell and hidden state updates and sigmoid activation for gates. Fully connected layers follow standard dense layer configurations. Weights are initialized using Glorot (Xavier) initialization {\cite{DL_FC.2010}}, and biases are zero-initialized.
All networks are trained using the Adam optimizer with an initial learning rate of 0.001 and gradient decay factor 0.9.
A piecewise learning rate schedule reduces the learning rate by a factor of 0.1 every 10 epochs.
L2 regularization with weight decay 0.0001 is applied to prevent overfitting.
Networks are trained for a maximum of 30 epochs with a mini-batch size of 50.
Validation is performed every 50 iterations with early stopping (patience = 5 epochs).
The model with the lowest validation loss is selected as the final network.}

\subsection{Single Association Network (SANT)}
Our association network uses a data-driven approach to solve the NP-hard data association problem {\cite{DA_hungarian.1955,DA_JPDA.1993}}, which traditionally requires significant computational effort for optimal solutions {\cite{DA_hungarian.1955}}.
Unlike conventional methods {\cite{DL_RNN_mot.2016, DL_RNN_data_association.2019}}, our \ac{SANT} model eliminates the need for a predefined distance matrix. Instead, it directly processes the current tracks and newly detected sensor objects, matching each new object to a track. This allows the network to autonomously learn its association strategy from training data, replacing the Hungarian Algorithm and a defined distance metric with a learning-based method. \red{Note that the incoming sensor objects $Z_{t,1:m}$ and the tracked objects $X_{t,1:n}$ already have the same state vector, no mapping is needed and the measurement equation of a traditional Kalman filter is reduced to a unit matrix \cite{welch2006introduction}.}

\begin{figure}[t]
	\centerline{\includegraphics[width=0.95\linewidth]{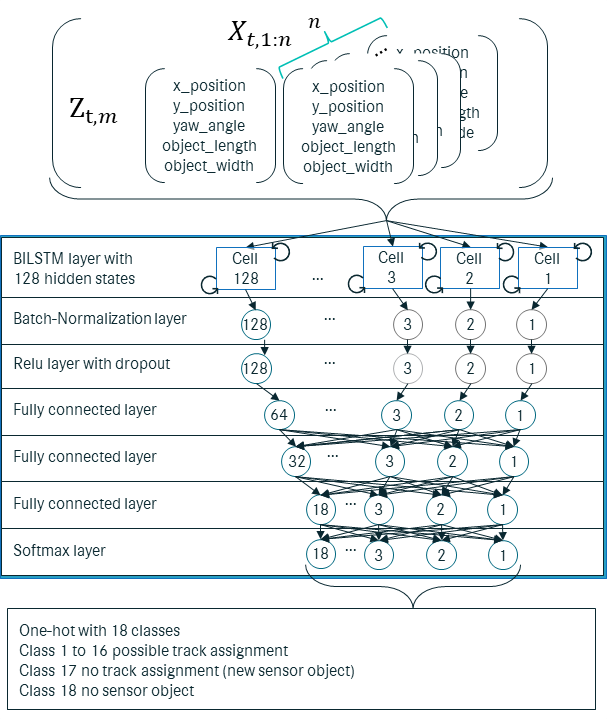}}
	\caption{Schematic representation of the network structure of \ac{SANT}. Here $m=1$, so one \ac{SO} is associated to $m$ existing tracks. \red{As input interface \ac{SANT} receives the predicted track states $X_{t,1:n}$ from \ac{SPENT} and one new sensor object $Z_{t,m}$ to be associated. The output is a one-hot vector indicating the association of the sensor object to one of the tracks or to none.}}
	\label{pic::SantArch}
\end{figure}

\paragraph{Data preprocessing}
In the formulation of \ac{SANT}, we conceptualized data association as a temporally structured challenge, adopting a sequence-to-vector paradigm.
In general, $m$ incoming sensor objects need to be associated to $n$ tracks.
For \ac{SANT} we focus on the association of a singular sensor object ($m=1$), denoted as $Z_{(t,m=1)}$, to $n$ tracks, represented as $X_{(t,1:n)}$.
These tracks were extracted from the KITTI dataset.
We note, however, that genuine hand-labelled \ac{GT} data for this specific association problem is not available. We rather use the existing tracks to  generate \red{synthetic} \ac{GT} data for the data association.
The input to \ac{SANT} consists of a single sensor object and a set of tracks (see top part of Fig. \ref{pic::SantArch}).
The output is a one-hot vector encoding the association of the incoming \ac{SO} to one of the tracks, or to none.
To generate the \red{synthetic} \ac{GT}, we take a set of tracks, then randomly chose one of them and take the next state vector at the next timestamp of that single track as the incoming new \ac{SO}. 
\red{To simulate measurement uncertainty while preserving a controlled training setup, we distort the ground-truth sensor object states with additive noise. Specifically, we apply zero-mean, independent Gaussian noise to each state component of the incoming sensor objects.
The noise magnitude is defined relative to the respective state value, with a maximum standard deviation of 3\% per dimension. This results in a scaled white Gaussian noise model that reflects typical relative inaccuracies observed in automotive perception systems, while avoiding unrealistic absolute perturbations for small state values.}

\red{The noise is applied independently per time step and per state dimension, assuming no temporal correlation and no cross-correlation between state variables. This design choice deliberately abstracts from detailed sensor-specific error models in favor of a lightweight and reproducible uncertainty approximation suitable for data-driven association learning.}
As shown in Fig. {\ref{pic::SantArch}}, the data format was created accordingly to enable index-based track assignment for \ac{SANT}.
The actual number of tracks can vary between 0 and a maximum of $n=16$ objects, as is given by the KITTI data set using our selected objects (cars and vans) {\cite{Geiger2012CVPR}}.
The size of the input matrix therefore corresponds to $k \times (n + 1)$, where $k = 5$ is the number of state values for our work.
The columns of the matrix \red{contain} the state vectors of the $n$ tracks and the state vector of the incoming sensor object.
A deeper analysis of our data is presented in the \ac{MANTa} section.

\paragraph{Network architecture}
The network as depicted in Fig. {\ref{pic::SantArch}} is designed as a sequence-to-classification network.
At each timestamp, a matrix is passed as input holding both, the information of the one to-be-assigned \ac{SO} and the multiple currently-tracked objects.
Architectures with \ac{RNN}, \ac{GRU}, \ac{LSTM} and \ac{BILSTM} layers were tested.
The best performing architecture was \red{experimentally} determined to be a \ac{BILSTM} layer as shown in Fig. {\ref{pic::SantArch}}.

We are using the work introduced by Hochreiter et al. {\cite{DL_LSTM.1997}}, who demonstrated the ability to capture the context from both ends of the sequence by combining the outputs of two LSTM layers that pass the information in opposite directions.
The resulting architecture is called \ac{BILSTM}.
The output mode has been configured in the \ac{BILSTM} layer, so that the layer is able to receive a sequence as input and calculate an output vector.
This form of dimension reduction is necessary in order to carry out the classification.
The final FC layer specifies the number of classes via the number of output values.
The class probabilities are calculated in the softmax layer by applying the softmax function.
The cross-entropy cost function is utilized to quantify the discrepancy between the network's probabilistic predictions and the ground truth values, a method particularly suited for tasks involving categorically exclusive classes.
This approach employs one-hot encoding to transform class representations into binary vector formats.

We calculate the cost function as the average of the cross-entropy losses for each prediction, relative to its corresponding target value:
\begin{equation}
    cost = -\frac{1}{N} \sum_{i=1}^{N} \sum_{j=1}^{C} y_{i,j} \log(\hat{y}_{i,j}),
\end{equation}
\noindent
\red{where $N$ denotes the} total number of association samples (i.e., total number of timestamps), $C$ represents the number of class categories, $y_{i,j}\in [0,1]$ is the GT indicator for whether class $j$ is the correct classification for sample $i$, and $\hat{y}_{i,j}$ is the predicted probability that sample $i$ belongs to class $j$, as derived from the softmax function output.

\subsection{Multi-Association Network (MANTa)}
The development of the \ac{SANT} demonstrated that data-driven association logic can be effectively learned by a deep learning model.
Building upon this foundation, we developed MANTa with the objective to create a network capable of associating multiple ($m$) sensor objects with multiple ($n$) tracks in a single operational step.
With MANTa, the following association scenarios can be addressed:

\begin{itemize}
    \item \textbf{1 to n} - one \ac{SO} to $n$ tracks 
    \item \textbf{m to 1} - $m$ \acp{SO} to one track
    \item \textbf{m to n} - $m$ \acp{SO} to $n$ tracks
    \item \textbf{m to 0} - $m$ \acp{SO} to no tracks
\end{itemize}

\begin{figure*}
	\includegraphics[width=0.95\linewidth]{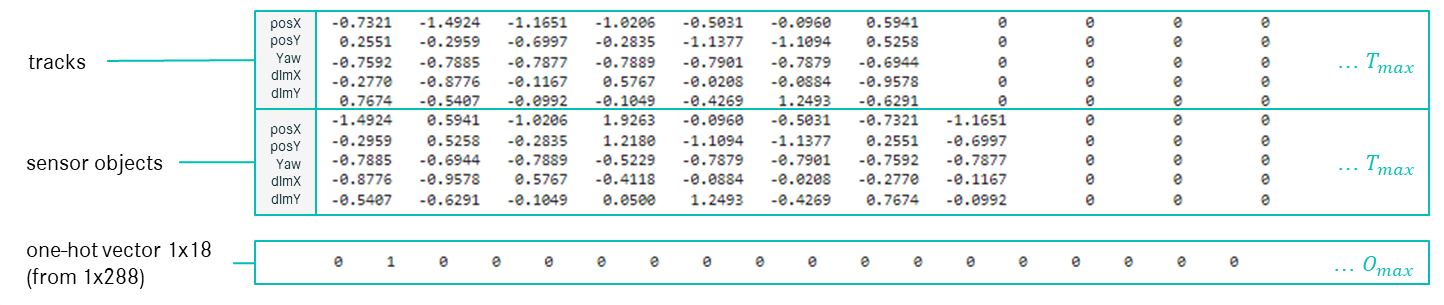}
	\captionof{figure}{MANTa, data structure, shows the non-noisy \acp{SO} to enable a visual assignment and increase understanding of the association procedure.
	Seven tracks are extracted from the KITTI data set for the given timestamp of sequence 20.
	Eight sensor objects are generated in pseudo-random order.
	The one-hot vector shows the GT assignment of the first sensor object to the track at position two.}
	\label{pic::MantaData}
\end{figure*}

\paragraph{Data preprocessing}
The association data set of \ac{MANTa} was created according to the described objective.
Fig. {\ref{pic::MantaData}} shows the input data structure with corresponding association tasks for timestamp 85 of sequence 20 of the KITTI data set. Equivalent data were extracted across all sequences.
All extracted tracks were each modified with noise as described in section \ref{ssec:spent} and then normalized.
Fig. {\ref{pic::MantaData}} shows an assignment example with non-noisy \acp{SO}. The association result is given by the one-hot vector at the bottom.
For each sensor object the network calculates a $1 \times 18$ one-hot vector containing its association result to the existing tracks.
For a maximum of $T_{\max}=16$ tracks, this results in an output vector of size $16 \times 18$.
The ordering of the \ac{SO}s is randomized per timestamp and the resulting association input matrix has dimension $F_{total} \times T_{\max}$, where $F_{total}$ represents the total number of features (here $2 \times k=10$, with $k$ being the number of values per state vector).
The one-hot vector depicted in Fig. {\ref{pic::MantaData}} shows the GT assignment of the first sensor object to the track at position two.
There are seven tracks in the timestamp of the sequence shown.
For each track a corresponding \ac{SO} is available.
Additionally, a new \ac{SO} was detected, leading to a total of eight sensor objects.
The \ac{GT} assignment of the first sensor object is shown in the one-hot vector at the bottom of Fig. {\ref{pic::MantaData}}.
The assignment output can be \red{thought of as a matrix} with dimensions maximum number of tracks $T_{\max}=16$ and the number of possible assignment classes $C=18$,
\red{where} each field can either be zero (no assignment) or one (assignment).
For numerical reasons, we unfold this matrix to a \red{vector of dimension $O_{\max}=288=16 \cdot 18$ with entries being 0 or 1}.

The assignment classes result from the described index class 1 to 16 and additional degrees of freedom.
One degree of freedom of the assignment represents the case that no measurement exists, another that the measurement should not be assigned.
As for SANT, the cross-entropy cost function calculates the cross-entropy loss between network predictions $\hat{y}_{i,j}$ and target values \red{$y_{i,j}$} for the unique assignment task for mutually exclusive classes.
\red{The introduced $O_{\max}=288$-dimensional vector} is used to represent the class in binary form.
The following formula is used to calculate the cross-entropy loss values for each timestamp $t$:

\begin{equation}
    loss_t = - \sum_{i=1}^{T_{\max}}  \sum_{j=1}^{C} \left[ y_{i,j} \ln(\hat{y}_{i,j}) + (1 - y_{i,j}) \ln(1 - \hat{y}_{i,j}) \right].
\end{equation}

\noindent
\red{Note that the elements of the one-hot vectors $\hat{y}_{i, j} \in \{0,1\}$ denote if track $i$ is associated to \ac{SO} $j $. Then, all scalars} obtained per timestamp are summarized and divided by the number of samples $N$ of a minibatch for the cost function:

\begin{equation}
    cost = \frac{1}{N} \sum_{t=1}^{N} loss_t.
\end{equation}

\begin{figure}[t!]
	\centerline{\includegraphics[width=0.95\linewidth]{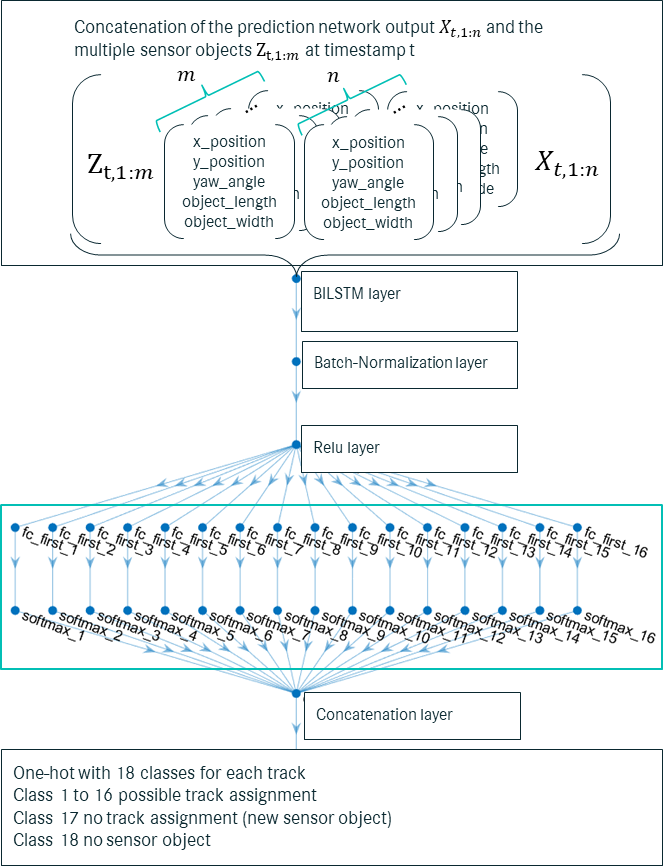}}
	\caption{Schematic representation of the generic network structure of MANTa.}
	\label{pic::MantaArch}
\end{figure}

\paragraph{Network architecture}
The schematic representation Fig. {\ref{pic::MantaArch}} shows the developed network architecture for the simultaneous association of a large number of sensor objects to a large number of tracks.
This is what we call a \ac{MANTa}.

The \ac{BILSTM} layer processes the input data as already explained for \ac{SANT}\@.
The task of associating a \ac{SO} list with a track list requires a separate network part for each track.
This extension is labelled accordingly in Fig. {\ref{pic::MantaArch}}.

For each track (from 1 to $T_{\max}=16$), the MANTa has been developed with the fully connected, softmax stack that was introduced for \ac{SANT}.
Each softmax output consists of a vector with $C=18$ elements, which represents the most probable assignment.
This means that a single assignment can be realized for each track.
The vectors 1 to $T_{\max}$ are linked together in the Concatenation Layer.
This creates a vector with 288 elements, whereby 18 elements each represent the most probable assignment of a \ac{SO} to a track.

\red{To summarize the operational flow of the proposed \ac{TbD} framework and to support reproducibility, algorithm {\ref{tab::pseudoalgo}} provides a concise pseudo-code representation of the overall tracking cycle integrating \ac{SPENT}, \ac{SANT}, and \ac{MANTa}.}

\begin{algorithm}[h]
    \caption{\red{Tracking-by-Detection Cycle with SPENT, SANT, and MANTa}}
    \label{tab::pseudoalgo}
    \begin{algorithmic}[1]
        \REQUIRE \red{Tracks ${T}_{t,1:n}$, sensor objects ${Z}_{t,1:m}$}
        \ENSURE \red{Confirmed tracks ${T^c}_{t,1:n}$}
        \STATE \red{Predict all track states ${X}_{t,1:n}$ using SPENT}
            \IF{\red{SANT is used}}
                \STATE \red{Iteratively associate sensor objects ${Z}_{t,1:m}$ to confirmed tracks ${T^c}_{t,1:n}$ to receive ${A}_{t,1:m}$}
            \ELSIF{\red{MANTa is used}}
                \STATE \red{Jointly associate sensor objects ${Z}_{t,1:m}$ to confirmed tracks ${T^c}_{t,1:n}$ to receive ${A}_{t,1:m}$}
            \ENDIF
        \STATE \red{Update associated tracks ${T^u}_{t,1:n}$ using ${A}_{t,1:m}$}
        \STATE \red{Perform track confirmation and deletion}
        \RETURN \red{${T^u}_{t,1:n}$}
    \end{algorithmic}
\end{algorithm}

\red{The pseudo-code explicitly distinguishes between iterative single-measurement association with \ac{SANT} and joint multi-measurement association with \ac{MANTa}, reflecting the framework-level selection strategy already described conceptually in Fig. {\ref{pic::MotFramework}}.}

\section{Experimental Evaluation}
The developed networks were modularly integrated into the Tracking-by-Detection framework, replacing classical algorithms.
\red{On a desktop CPU system without AI acceleration hardware (Intel Core i7-13850HX Intel64, single-threaded execution),
median inference times were measured in FP32 precision.
SPENT requires approximately 0.5 ms per track, resulting in 8.0 ms for 16 concurrent tracks.
SANT requires approximately 1.0 ms per association for $n\leq16$ tracks, while MANTa performs a joint multi-object association in approximately 3.5 ms total. For a typical Tracking-by-Detection cycle with up to 16 tracks, this corresponds to an average processing time of approximately 0.2 ms per track when normalized over all tracks.}

\begin{figure}[h!]
	\centerline{\includegraphics[width=0.87\linewidth]{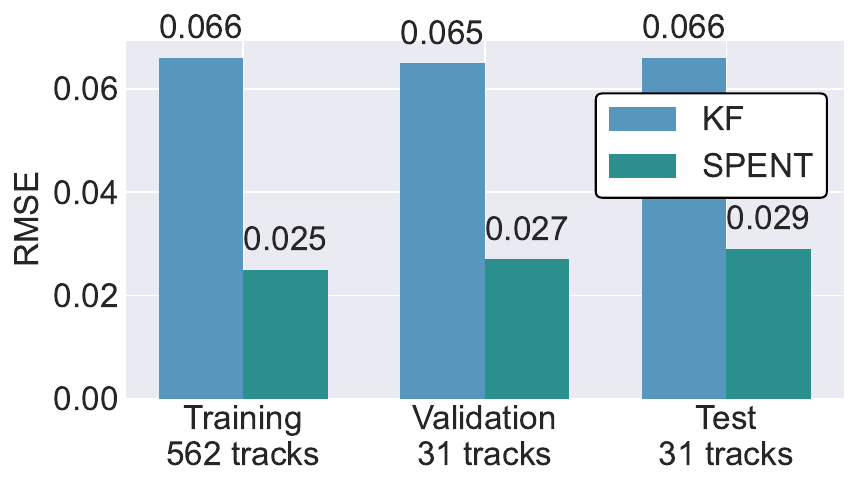}}
    \caption{\red{Comparison of the \ac{RMSE} for \ac{SPENT} and a \ac{KF} framework implemented by the Daimler Truck Research Group following {\cite{KF_BoundingBD.2023,KF_Bewley.2016}}.}}
    \label{fig::SpentRmseComparison}
\end{figure}

Our recurrent network updates its internal hidden states at each timestamp, capturing temporal dependencies and enabling accurate state predictions without external correction.
This approach, well-suited for real-time applications presents a strong alternative to traditional methods.
Using the KITTI dataset, focusing on vehicle tracks (cars and vans) divided into training, validation, and testing sets, we evaluated \ac{SPENT}'s performance.
As first benchmark, we take a \ac{KF} framework implemented by the Daimler Truck Research Group following {\cite{KF_BoundingBD.2023,KF_Bewley.2016}}, which achieves a RMSE of 0.066 across 31 tracks (Fig. {\ref{fig::SpentRmseComparison}}) on the testing set.
Our \ac{SPENT} model reduces the RMSE by more than half to 0.029 using the identical data sets. For positional predictions the average deviations \red{are} 42 centimeters on the x-axis and 23 centimeters on the y-axis.

For the association of sensor objects to existing tracks, we developed \ac{SANT} to replace classical methods like \ac{GNN}. By substituting the distance metric and the Hungarian assignment procedure with a learned, data-driven assignment logic, \ac{SANT} achieves an accuracy of 95\% on a testing dataset with 391 samples (representing 5\% of the total dataset with 7827 association samples). This approach not only simplifies the assignment process but also allows for adaptability in diverse scenarios, where classical methods may lack flexibility.
\red{When positioning our approach relative to detection‑centric or transformer‑based trackers, it is worth noting that methods such as Tracktor, TransTrack and TrackFormer pursue alternative design points (detector‑based heuristics or end‑to‑end temporal modeling) which trade modularity for different operational benefits {\cite{Bergmann2019Tracktor,Sun2020TransTrack,Meinhardt2021TrackFormer}}.}

\begin{figure}[b!]
	\centerline{\includegraphics[width=0.87\linewidth]{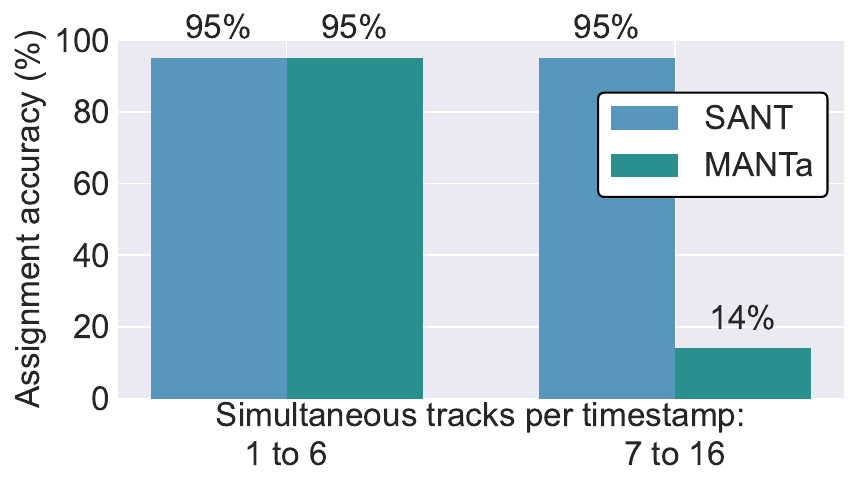}}
    \caption{\red{Data association results for the SANT and MANTa networks. MANTa's average accuracy is 80\%.}}
    \label{fig::SANTMANTaAccuracy}
\end{figure}

Expanding on \ac{SANT}, \ac{MANTa} addresses the limitations of a single object to track assignment by enabling multi object assignments within each timestamp.
This means \ac{MANTa} is trained to assign a list of \acp{SO} to a list of tracks in a single operational step on the same dataset as \ac{SANT} and also achieved an assignment accuracy of 95\% for the six most frequently occurring association sets, i.e., the sets with one to six tracks per timestamp. It performs much worse (14\%) for assignment scenarios with more tracks, which were less present in the training data. Accuracy results are summarized in Fig. \ref{fig::SANTMANTaAccuracy}.

\begin{figure}[t!]
    \centerline{\includegraphics[width=0.95\linewidth]{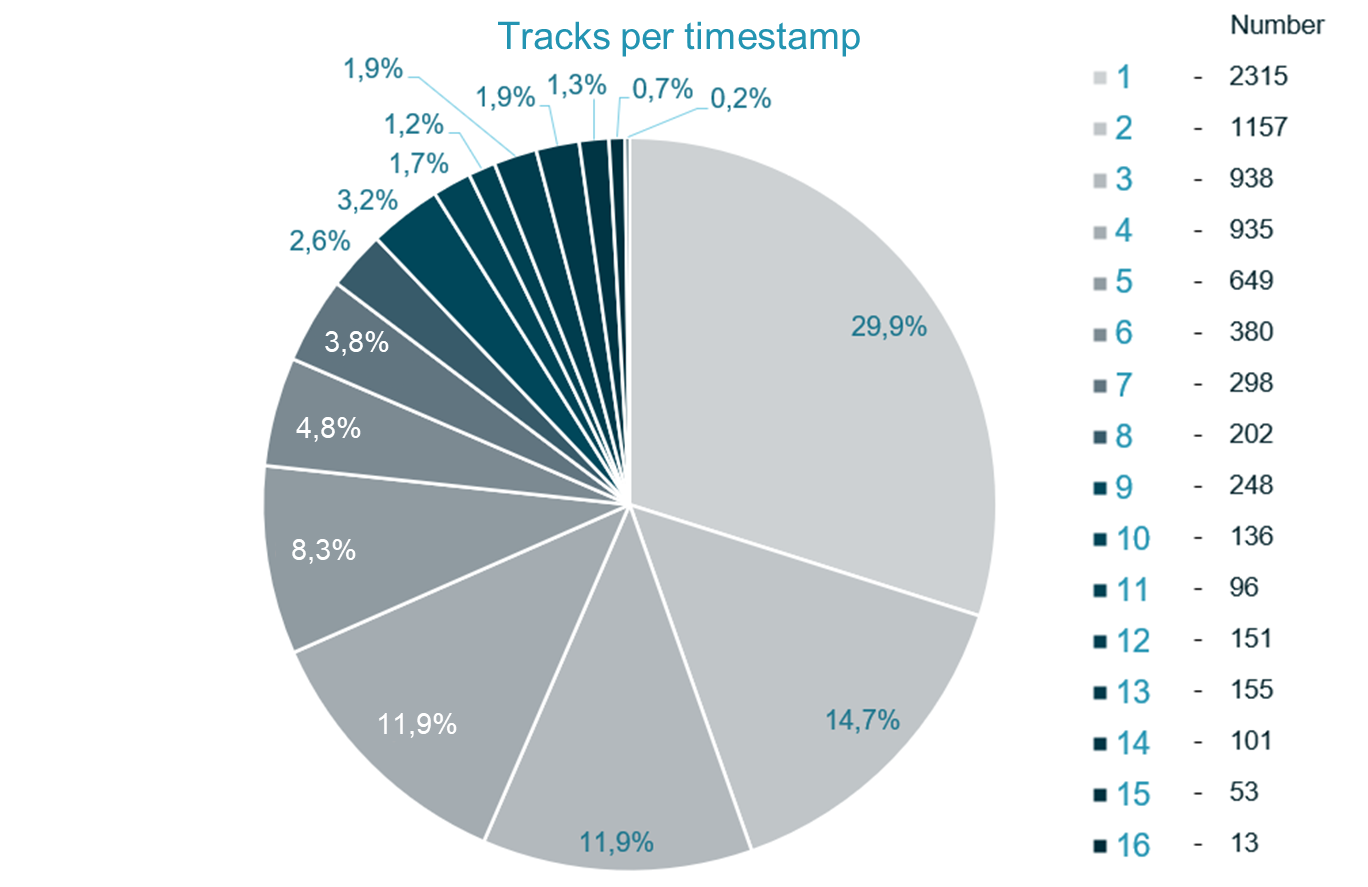}}
	\caption{The diagram shows the distribution of the number of existing tracks per timestamp.
		For the KITTI dataset, which includes both car and van objects. It reveals that timestamps containing one to six tracks constitute 81.5\% of the samples (6374 of 7827).
		The legend shows the absolute number of samples which contain the respective number of tracks.
		In 29.9\% of the samples, the data contains one track, in 14,7\% two tracks. Only 18.5\% (1448 of 7827 samples) of the samples contain more than six tracks, which leads to an unbalanced dataset and makes learning the association for \ac{MANTa} for 6+ tracks harder.}
	\label{pic::MantaTracks}
\end{figure}

In the context of the entire KITTI dataset, which includes both cars and vans objects,
MANTa achieves an average association accuracy of 80\%.
This polarized performance with limitations for seven or more tracks was primarily attributed to the characteristics of the extracted data.
As illustrated in Fig. {\ref{pic::MantaTracks}}, the distribution of the number of existing tracks per timestamp reveals significant insights.

Notably, timestamps containing exactly one track constitute nearly one-third of the entire dataset, accounting for 29.9\% of the data, which corresponds to an absolute count of 2315 samples.
Timestamps containing one to six tracks constitute 81.5\% of the samples and are therefore 6374 of 7827 samples.
Consequently, tests were conducted using a reduced dataset with one to six tracks per timestamp to demonstrate the multi-association capability of the network.
\ac{MANTa} correctly assigns 95\% of the dataset for timestamps containing one to six tracks.
This result points to MANTa's proficiency in handling data given the appropriate dataset, as SANT also achieves a validation accuracy of approximately 95\% across the entire KITTI dataset (including cars and vans).
The primary advantage of \ac{MANTa} over \ac{SANT} is its ability to assign multiple sensor objects to multiple tracks in a single operational step.

\section{Conclusion and Future Work}
\red{The core philosophy of our work was to improve a Kalman tracking framework for \ac{ADAS} using machine learning techniques, while maintaining modularity and interpretability—two core strengths of these classical approaches. To this end,} we designed, trained and evaluated three novel \ac{NN}s: (i) SPENT for the prediction of tracked objects, (ii) SANT for the association of one new \ac{SO} to a list of tracks and (iii) MANTa for the association of multiple SOs to a list of tracked objects. All networks were developed for real-time embedded applications with none having more than 50k trainable parameters.
Our approach leaves the general structure of a \ac{KF} framework intact, preserving modularity, interpretability and the ability to test each component separately. This work lays a foundation for future \ac{ADAS} research, highlighting the potential of data-driven software development in overcoming the limitations of classical algorithms, which in practice often need to include heuristics.

Future research topics will include: (a) \red{while the experimental evaluation is conducted on the KITTI dataset, which is a standard benchmark for automotive \ac{MOT}, future work will extend the validation to additional datasets to further assess transferability across different sensor configurations and traffic scenarios}, (b) the quantification of uncertainties to improve decision making and (c) the investigation of multitasking networks to optimize the context-dependent tracking of multiple objects.

\section*{Declarations}
\small{
\noindent
\textbf{Abbreviations:} ADAS, Advanced Driver Assistance Systems; BILSTM, Bidirectional Long Short-Term Memory; FC, Fully Connected; GNN, Global Nearest Neighbor; GRU, Gated Recurrent Unit; GT, Ground Truth; HA, Hungarian Algorithm; JPDA, Joint Probabilistic Data Association; KF, Kalman Filter; LSTM, Long Short-Term Memory; MANTa, Multi-Association Network; MSE, Mean Squared Error; ML, Machine Learning; MOT, Multi-Object Tracking; NN, Neural Network; RMSE, Root Mean Square Error; RNN, Recurrent Neural Network; SANT, Single-Association Network; SO, Sensor Object; SoDA, Soft Data Association; SPENT, Single-Prediction Network; TbD, Tracking-by-Detection.

\noindent
	\textbf{Availability of data and materials:} All used datasets are publicly available, see {\cite{Geiger2012CVPR}} for details. Requests for model parameters can be \red{sent} to the authors. Since the models were developed within Daimler Truck AG internal research, no direct download is provided.

\noindent
\textbf{Competing interests:} All authors declare that there are no competing interests.

\noindent
\textbf{Funding:} Not applicable.

\noindent
\textbf{Acknowledgments:} Not applicable.

\noindent
	\textbf{\red{Authors' contribution}:} All authors contributed to the study’s conception and design. CH and CB performed the implementations. CH performed the data preparation and experiments. The first draft of the manuscript was written by CH and MD. All authors commented on and extended the first version of the manuscript. All authors read and approved the final manuscript.
}

\bibliographystyle{IEEEtran}
\bibliography{IEEEabrv, lit.bib}

\vfill

\end{document}